\documentclass[letterpaper]{article} 
\usepackage{aaai25}  
\usepackage{times}  
\usepackage{helvet}  
\usepackage{courier}  
\usepackage[hyphens]{url}  
\usepackage{graphicx} 
\usepackage{natbib}  
\usepackage{caption} 
\usepackage{amsmath}
\usepackage{algorithm}
\usepackage{algpseudocode}

\usepackage{url}            
\usepackage{booktabs}       
\usepackage{amsfonts}       
\usepackage{nicefrac}       
\usepackage{microtype}      
\usepackage{xcolor}
\usepackage{times}
\usepackage{soul}
\usepackage{graphicx}
\usepackage{amsmath}
\usepackage{amsthm}
\usepackage{amssymb}
\usepackage{mathtools}
\usepackage{bm} 
\usepackage{subfigure}
\usepackage{multirow}
\usepackage{bbding}
\usepackage{enumitem}

\usepackage[textsize=tiny]{todonotes}
\usepackage{longtable}
\usepackage[capitalize,noabbrev]{cleveref}
\usepackage{algorithm}
\usepackage{algpseudocode}

\usepackage{newfloat}
\usepackage{listings}
\DeclareCaptionStyle{ruled}{labelfont=normalfont,labelsep=colon,strut=off} 
\floatstyle{ruled}
\newfloat{listing}{tb}{lst}{}
\floatname{listing}{Listing}
\title{Enhancing LLM Agents for Code Generation with Possibility and Pass-rate Prioritized Experience Replay}
\author{
    Yuyang Chen\textsuperscript{$\dagger$1,2},
    Kaiyan Zhao\textsuperscript{$\dagger$1},
    Yiming Wang\textsuperscript{3}, 
    Ming Yang\textsuperscript{3},
    Jian Zhang\textsuperscript{1},
    Xiaoguang Niu\textsuperscript{1}
}
\affiliations{
    \textsuperscript{1}School of Computer Science, Wuhan University, Wuhan, China\\
    \textsuperscript{2}Northwestern University, Evanston, IL, USA\\
    \textsuperscript{3}State Key Laboratory of Internet of Things for Smart City, University of Macau, Macao, China\\
    chenyuyang0520@gmail.com, \{zhao.kaiyan, jzhang, xgniu\}@whu.edu.cn, \{wang.yiming, ming.ink\}@connect.um.edu.mo
}

\usepackage{bibentry}

\begin{document}

\maketitle

\begin{abstract}
Nowadays transformer-based Large Language Models (LLM) for code generation tasks usually apply sampling and filtering pipelines. Due to the sparse reward problem in code generation tasks caused by one-token incorrectness, transformer-based models will sample redundant programs till they find a correct one, leading to low efficiency. To overcome the challenge, we incorporate Experience Replay (ER) in the fine-tuning phase, where codes and programs produced are stored and will be replayed to give the LLM agent a chance to learn from past experiences. Based on the spirit of ER, we introduce a novel approach called BTP pipeline which consists of three phases: beam search sampling, testing phase, and prioritized experience replay phase. The approach makes use of failed programs collected by code models and replays programs with high Possibility and Pass-rate Prioritized value (P2Value) from the replay buffer to improve efficiency. P2Value comprehensively considers the possibility of transformers' output and pass rate and can make use of the redundant resources caused by the problem that most programs collected by LLMs fail to pass any tests. We empirically apply our approach in several LLMs, demonstrating that it enhances their performance in code generation tasks and surpasses existing baselines.

\end{abstract}
\section{Introduction}

In recent years, there has been significant progress in the development of Large Language Models (LLMs) like Transformer \cite{vaswani2017attention} and Llama \cite{touvron2023llama} across various domains. A particular trend has emerged in leveraging LLMs for automatic code generation tasks. Models such as WizardCode \cite{luo2023wizardcoder} and StarCode \cite{li2023starcoder} have been developed to address these tasks. To evaluate the effectiveness and performance of LLMs in code generation, various benchmarks have been established. For instance, APPS \cite{hendrycks2021measuring} is widely used in evaluations of code models, and Code-Contests \cite{li2022alphacode} has been established as a standard for competition-level coding tasks. Among all models, transformers have demonstrated significant success in benchmarking tasks such as code translation, code completion, and challenging problem-solving \citep{svyatkovskiy2020code}. Some transformer-based pipelines have even achieved remarkable results on difficult tasks \citep{zhang2023planning}.

However, traditional transformer-based pipelines, which consist of sampling and filtering phases, have obvious shortcomings due to their structure. A salient problem in code generation tasks is the significant waste of redundant resources caused by low efficiency. Specifically, when tasks are provided as input, the code agent samples a large number of programs from pre-trained transformer-based LLMs and passes them through public test sets, where they are tested and filtered based on their pass rates. Low efficiency arises in cases where most programs fail to pass the tests due to even a single incorrect token \citep{zhang2023planning}. As a result, models must sample many incorrect programs to find a precisely accurate one, leading to the wastage of redundant resources, including unsuccessful programs that are not reused.

However, programs that fail some tests do not necessarily lack value. On the contrary, most pre-trained LLMs are well-trained on large corpora, which means that the programs they generate are almost accurate but may fail due to minor errors. Thus, it would save time and improve efficiency if we could reduce the waste of these valuable resources.

To leverage the value hidden in these redundant resources and increase efficiency, we introduce Experience Replay (ER), a buffer that stores programs sampled by LLMs along with each program's P2Value (possibility and pass rate value), which we consider as its value. P2Value comprehensively considers both the likelihood of a transformer's output and the pass rate. On one hand, a program with a higher pass rate in public test sets demonstrates good performance in a particular task; on the other hand, a program with a higher likelihood of output is considered to have higher value according to the results calculated by pre-trained transformers. Based on ER, we introduce a novel approach called the BTP pipeline, which consists of three phases: beam search sampling, testing, and prioritized experience replay. The core algorithm, PPER (P2Value-Prioritized Experience Replay), utilizes beam search to sample and store programs in ER, and then replays programs in ER based on a probability dependent on their P2Value.

We empirically demonstrate that our pipeline improves the performance of LLMs in code generation tasks, outperforming the original models regardless of whether the training data is self-generated or generated by models of higher quality. More specifically, our contributions are as follows:
\begin{itemize}
\item First, we propose a novel algorithm, \textbf{BTP pipeline}, consisting of beam search sampling phase testing phase and experience replay phase to fine-tune LLMs. 
\item We empirically demonstrate that our algorithm performs well not only in scenarios where better LLMs generate data to fine-tune normal LLMs, but also in scenarios where LLMs sample programs to fine-tune themselves. That is to say, our BTP pipeline is generic.
\item At last, we introduce a novel buffer called experience replay buffer which is efficient in fine-tuning. In the future, we can combine it with other algorithms to find more efficient way to fine-tune LLMs.
\end{itemize}
\section{Related Work}

Considering the association of ER and LLMs in our pipeline, here we will introduce them respectively.

\textbf{LLMs for code generation:} Our work is closely related to LLMs for code generation. In recent years, high-performing LLMs such as GPT-4 \cite{gpt42023technical}, Llama \cite{touvron2023llama}, PaLM \cite{chowdhery2022palm}, and Chinchilla \cite{hoffmann2022chinchilla} have emerged in different areas. Particularly, our work is based on transformers \cite{vaswani2017attention} for code generation tasks \cite{roziere2020transformers}. Code models such as BERT \cite{devlin2019bert,feng2020codebert,guo2020graphcodebert}, T5 \cite{raffel2020exploring}, GPT-2 \cite{radford2019language}, Codex \cite{chen2021codex}, CodeT5 \cite{wang2021codet5}, StarCode \cite{li2023starcoder}, and WizardCoder \cite{luo2023wizardcoder} have become backbones for code understanding and generation. Meanwhile, to evaluate the performance of code models, different benchmarks have been created, such as APPS \cite{hendrycks2021measuring}, CODE-CONTESTS \cite{li2022alphacode}, OpenOrca \cite{lian2023openorca}, and HumanEval \cite{chen2021evaluating}. Additionally, \citep{roziere2022unittest} constructs training datasets for unsupervised code translation tasks, and \citep{ellis2019synthesizing} constructs test cases from different specific areas to train an RL agent. Moreover, many new methods have been proposed in code generation. For example, \citep{chen2021evaluating} fine-tunes powerful pre-trained LLMs to index knowledge and refine performance in code completion. \citep{austin2022program} summarizes that LLMs can be applied to code generation tasks, and \citep{wei2022chain} introduces Chain-of-Thought (CoT) prompting to encourage LLMs to think step by step and reduce error rates.

\textbf{RL for Code Generation:} \citep{bunel2018leveraging} claims that code generation tasks can be broken down into a series of decision-making problems, which are similar to problem definitions in RL. This implies that RL algorithms can be applied to transformers, effectively leveraging their sequential decision-making capabilities. For instance, \citep{zhang2023planning} combine Monte Carlo Tree Search (MCTS) with transformers in code generation tasks. In this approach, MCTS is used to explore potential sequences of code by simulating different code paths, evaluating them based on a reward function that measures code correctness and efficiency. This enables the model to select the most promising code sequences during inference\citep{yangming1}. Similarly, \citep{le2022reinforcement} optimize the correctness of generated programs by framing it as a reward maximization problem, a common objective in RL\citep{yangming2}. They employ policy gradient methods, where the transformer model's policy is iteratively improved by sampling code sequences, estimating the reward (e.g., program correctness)\cite{yiming1,yiming2}, and updating the model to increase the likelihood of generating correct code sequences in future iterations. These approaches demonstrate how RL's exploration-exploitation\citep{yangming3} mechanisms can be integrated with transformers to enhance the performance of code generation models by not just predicting the next token but optimizing over entire sequences based on cumulative rewards.

\textbf{Experience Replay:} \citep{lin1992self,kaiyan1} first introduced the concept of experience replay, suggesting that an agent can store its experiences in a buffer and later sample from this buffer to break the temporal correlation of consecutive observations, thus stabilizing the learning process. By replaying these experiences multiple times, the agent can improve sample efficiency, as it can learn from past experiences that may have been missed during the initial training phase.

Hindsight Experience Replay (HER) \citep{andrychowicz2017hindsight} expanded on this by addressing the sparse reward problem in goal-conditioned reinforcement learning (RL). In HER, after a failed episode, the transitions leading up to the failure are stored in the replay buffer. During training, these transitions are re-labeled with different goals than originally intended, particularly goals that were actually achieved in the failed episode. By assigning new rewards corresponding to these new goals, HER enables the agent to learn from episodes where the original goal was not achieved, effectively turning failures into learning opportunities.

Meanwhile, Prioritized Experience Replay (PER) \citep{schaul2015prioritized} introduced the idea that not all transitions are equally valuable for learning. PER modifies the experience replay mechanism by sampling transitions with a probability proportional to their temporal-difference (TD) error, which indicates how surprising or unexpected the transition is. High TD-error transitions, where the agent's prediction was significantly different from the actual outcome, are more informative and thus are replayed more frequently. This approach ensures that the agent focuses on learning from the most informative experiences, speeding up the convergence of the learning process by reducing the time spent on less useful transitions.

\section{Methodology}

\begin{figure*}[!htb]
\centering
\includegraphics[width=0.8\textwidth]{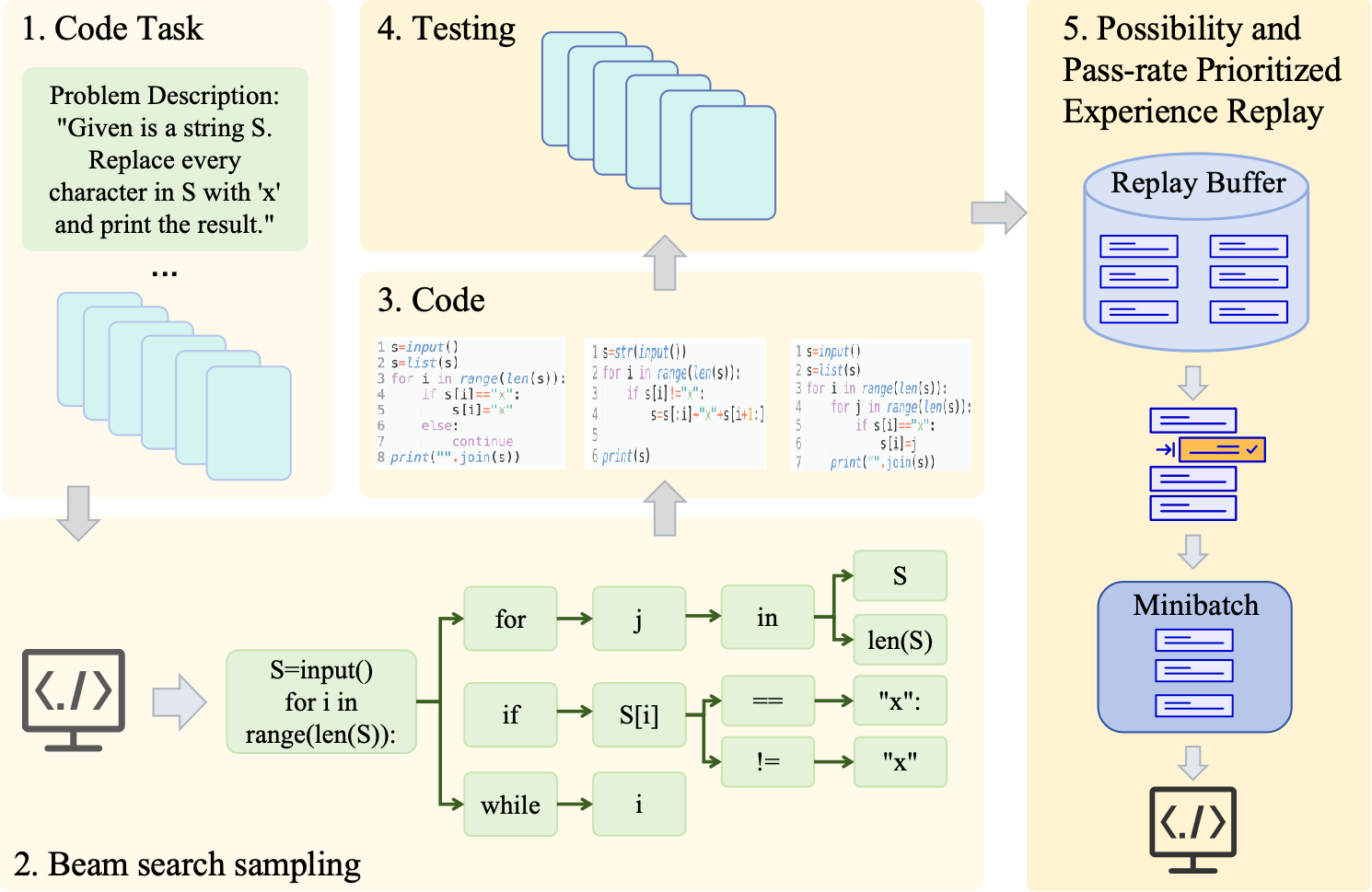}
\caption{The pass rate of GPT2-Wizard using BTP in different $\alpha$}
\label{fig:example}
\end{figure*}

In this section, we first briefly present the framework of the BTP pipeline, which is composed of beam search sampling, testing, and PPER. Then, we illustrate the details of the proposed framework. Figure 1 provides a complete process of the BTP pipeline.

\subsection{BTP pipeline}

As shown in Figure 2, most previous works adopt a traditional framework where the transformer model utilizes a sampling and filtering pipeline. In the sampling phase, LLM simply finds best code sequences in every time step according to 

\begin{equation}
\begin{aligned}
P(Y|X) &= P(y_1|X) P(y_2|X, y_1) \dots \\
&\quad P(y_T|X, y_1, y_2, \dots, y_{T-1})
\end{aligned}
\end{equation}

where X is the input, $y_i$ is the token generated in time step i, $P(y_i|X, y_1, y_2, \dots, y_{i-1})$ is the conditional probability of $y_i$ given X, $y_1$,...,$y_{i-1}$.

Particularly, in code generation tasks, X is a code task in the formulation of prompt.Traditional LLM simply uses greedy algorithm and maximizes $P(y_i|X, y_1, y_2, \dots, y_{i-1})$ for every time step i till gets a complete code sequence $y_1y_2\dots y_T$

In the filtering phase, the code sequence is sent to public test sets. However, it may fail due to some token mistakes. One main reason is that the greedy algorithm ignores the values hidden in other sequences with lower probabilities. Therefore, we introduce beam search sampling in our pipeline.
\subsection{Beam search sampling phase}
As shown in Figure 4, at every time step, the code model finds the top-k most probable candidate sequences and keeps them in a container called "beams." At each step i, all candidate sequences explore all possible tokens from the vocabulary, generating different new sequences. Then, the LLM selects the top-k sequences based on the combined probability and adds them to the new beam for the next time step.
\begin{equation}
P(y_1y_2\dots y_{i})=P(y_1y_2\dots y_{i-1})P(y_{i})
\end{equation}
Repeat the process till the model finds a complete program $y_1y_2\dots y_{T}$where T is the end time step. Store the top-k programs $t_1t_2\dots t_k$ in the experience replay buffer along with their possibilities $P(t_1)P(t_2)\dots P(t_k)$, the code task X and its corresponding test sets S in the following tuple form:

\begin{equation}
T=(X,S, t_i,P(t_i))
\end{equation}
\subsection{Testing phase}
In the testing phase, the model will sequentially take out every tuple from T and test $t_i$ in every test case of test set S, and compute the pass rate $pass\_rate_i$:
\begin{equation}
\text{pass\_rate}_{i} = \sum_{for S_k \in S} \mathbf{1}(\text{if } t_{i} \text{ pass } S_k)
\end{equation}
The pass rate, together with the combined probability, will be stored in the experience replay buffer (ER).
\begin{equation}
ER_i=(X,S, t_i,P(t_i),pass\_rate_i)
\end{equation}

\subsection{PPER phase}
In the Possibility and Pass-rate Prioritized Experience Replay (PPER) phase, the code model will be fine-tuned using the method we call PPER. Specifically, the programs stored in the replay buffer will be sampled with probabilities that are associated with their possibility and pass rate. The sampled programs will then be used to construct a minibatch, which will be used to fine-tune the code model.

\subsubsection{P2Value}
In our PPER method, the most important factor is to establish a standard for defining the priorities of every program in the ER. While it is challenging to determine an accurate measurement standard for priority, a reasonable alternative is to consider the P2Value, which combines the output probability of the transformer and the pass rate on test sets. Particularly for any tuple $ER_i=(X,S, t_i,P(t_i),pass\_rate_i)$ sampled from ER, P2value is calculated as followed:
\begin{equation}
\text{P2Value} = \alpha \cdot P(t_i) + (1 - \alpha) \cdot pass\_rate_i
\end{equation}
Where $\alpha$ is a parameter that determines the weights of possibility and pass rate.The closer it gets to 1, the more important the possibility becomes. Correspondingly, sampled programs that the original code model prefers will have more influence in the fine-tuning process. Conversely, programs that pass the most test sets but are not as highly preferred by the original code model will carry more weight in the fine-tuning process.

The reason we consider such a formula is that programs with higher pass rates are more suitable and valuable for particular code tasks. However, due to the possibility of low pass rates, which can even approach zero, we consider applying possibility to value a program. It is evident that a program preferred by the pre-trained LLM holds higher value in the LLM's corpus. And how to balance their weights is the reason why we set parameter $\alpha$
\subsubsection{Random Proprotization sampling}
It is straightforward to uniformly sample programs from the ER or to fine-tune the LLM using the entire ER. However, it is more efficient to give programs with higher value a greater chance of being selected. Therefore, we introduce a random sampling method to ensure that every program stored in the ER is sampled in a strictly monotonic manner with respect to its priority. This method increases the likelihood of sampling programs with higher priority, while still maintaining a fixed non-zero probability of sampling the program with the lowest priority, ensuring that every trajectory in the ER is utilized.

Specifically, we define the sampling probability of a transition i in Equation 5.
\begin{equation}
P(i) = \frac{p_i^\alpha}{\sum_k p_k^\alpha}
\end{equation}
 where pi is the priority of program $t_i$. The index $\alpha$ determines the level of prioritization, with $\alpha$ = 0 
 We consider two ways to define pi.
 In the first case, we directly define $p_i$ as P2Value. It intuitively depicts the relationship between sampling possibility and priority.

However, this method is sensitive to points that deviate significantly from the average value. For instance, trajectories with much higher P2Value will be sampled too frequently.

To solve this problem, we introduce the second definition
 \begin{equation}
p_i = \frac{1}{\text{rank}(i)}
\end{equation}
where rank(i) represents the rank of the program's priority among all trajectories. This method has several advantages compared with the previous approach. Firstly, it follows a power law distribution, meaning that most data are concentrated around the centroid, while a small proportion is distributed around the very large and very small values. Moreover, it is more robust and less sensitive to points that deviate significantly from the average value. For instance, P(i) of a trajectory with the lowest rank will not vary significantly even if its P2Value decreases substantially.

It is noteworthy that the possibility of the code model's output is non-zero, which means that P2Value is also non-zero, so there is no need for a constant to prevent zero probability.

\begin{algorithm}[H]
\caption{BTP Pipeline}
\label{alg:BTP}
\begin{algorithmic}[1]
\Require $T$: Code model; $beam$: a buffer that stores programs in the beam search sampling phase; $k$: size of beam; $X_{\text{set}}$: task sets with test sets; $ER$: experience replay buffer; $batch$: a minibatch that stockpiles programs used to fine-tune $T$; $n$: size of minibatch
\State \textbf{Beam Search Sampling Phase}
\For{each $(X, S)$ in $X_{\text{set}}$}
    \State $beam \gets \text{Beam\_search}(X, k)$
    \For{each $(t, P)$ in $beam$}
        \State Store $(X, S, t, P)$ in $ER$
    \EndFor
\EndFor

\State \textbf{Test Phase}
\For{each $(X, S, t, P)$ in $ER$}
    \State Test $t$ in $S$ and get $pass\_rate$
    \State Replace $(X, S, t, P)$ with $(X, S, t, P, pass\_rate)$
\EndFor

\State \textbf{PPER Phase}
\For{$i \gets 1$ to $n$}
    \State Sample $(X, S, t, P, pass\_rate)$ from $ER$ with probability according to $\text{P2Value}$
    \State Store $(X, S, t)$ in $batch$
\EndFor
\State Fine-tune $T$ with $batch$
\end{algorithmic}
\end{algorithm}
\section{Experiments}
\begin{table*}[!htb]
\centering
\renewcommand{\arraystretch}{1.5} 
\scalebox{0.84}{
\begin{tabular}{lcccccccc}
\toprule
 & \multicolumn{4}{c}{\textbf{Pass Rate (\%)}} & \multicolumn{4}{c}{\textbf{Accuracy rate(\%)}} \\
\cmidrule(lr){2-5} \cmidrule(lr){6-9}
\textbf{} & \textbf{APPS Intro.} & \textbf{APPS Inter.} & \textbf{APPS comp.} & \textbf{APPS mixed} & \textbf{APPS Intro.} & \textbf{APPS Inter.} & \textbf{APPS comp.} & \textbf{APPS mixed} \\
\midrule
\textbf{APPS GPT-2} & & & & & & & & \\
\textbf{GPT-2} & 12.37 & 10.67 & 4.33 & 7.30 & 5.30 & 3.30 & 1.32 & 2.72 \\
\textbf{GPT-2-GPT4} & 50.79 & 43.27 & 37.68 & 40.91 & 25.84 & 19.87 & 15.34 & 20.21 \\
\textbf{GPT-2-GPT3.5} & 41.68 & 37.62 & 28.59 & 32.25 & 19.42 & 16.25 & 12.21 & 15.63 \\
\textbf{GPT-2-Llama} & 35.82 & 31.14 & 24.37 & 27.60 & 14.35 & 11.40 & 8.27 & 11.69 \\
\textbf{GPT-2-Wizard} & 33.76 & 29.08 & 22.31 & 25.54 & 12.29 & 9.34 & 6.21 & 9.63 \\
\midrule
\textbf{APPS GPT-Neo} & & & & & & & & \\
\textbf{GPT-Neo} & 14.32 & 9.80 & 6.39 & 5.73 & 6.70 & 2.00 & 2.10 & 3.31 \\
\textbf{GPT-Neo-GPT4} & 51.23 & 46.39 & 38.89 & 42.12 & 26.88 & 23.34 & 17.12 & 20.54 \\
\textbf{GPT-Neo-GPT3.5} & 39.23 & 34.39 & 26.89 & 30.12 & 14.88 & 11.34 & 5.12 & 8.54 \\
\textbf{GPT-Neo-Llama} & 38.24 & 33.40 & 25.90 & 29.13 & 13.89 & 10.35 & 4.13 & 7.55 \\
\textbf{GPT-Neo-Wizard} & 35.83 & 30.99 & 23.49 & 26.72 & 11.48 & 8.54 & 2.32 & 5.74 \\
\bottomrule
\end{tabular}}
\caption{Result of "Better models help fine-tune normal models" experiment. On the top and bottom of the table, we show the performance of GPT-2 and GPT-Neo, and how they perform after they are fine-tuned by programs sampled by better models including GPT-4-turbo, GPT-3.5-turbo, CodeLlama-
34B, WizardCoder-34B}
\label{table:accuracy1}
\end{table*}

In this section, we empirically measure the effectiveness of our BTP pipeline. We conduct experiments sequentially to verify the following conjectures.

\begin{table*}[!htb]
\centering
\renewcommand{\arraystretch}{1.5} 
\scalebox{0.84}{
\begin{tabular}{lcccccccc}
\toprule
  & \multicolumn{4}{c}{\textbf{Pass Rate (\%)}} & \multicolumn{4}{c}{\textbf{Accuracy rate(\%)}}\\
\cmidrule(lr){2-5} \cmidrule(lr){6-9}
\textbf{} & \textbf{APPS Intro.} & \textbf{APPS Inter.} & \textbf{APPS comp.} & \textbf{APPS mixed}& \textbf{APPS Intro.} & \textbf{APPS Inter.} & \textbf{APPS comp.} & \textbf{APPS mixed} \\
\midrule
\textbf{GPT-2} & 12.37 & 10.67 & 4.33 & 7.30 &5.30 &3.30 & 1.32 &2.72\\
\textbf{GPT-2-2} & 15.36 & 11.23 & 5.11 & 8.12 &4.25 &2.78 & 2.25 &3.92\\
\midrule
\textbf{GPT-Neo} & 14.32 & 9.80 & 6.39 & 5.73 &6.70 &2.00 & 2.10 &3.31\\
\textbf{GPT-Neo-Neo} & 14.02 & 9.23 & 7.25 & 5.39 &6.32 &2.13 & 2.92 &3.72 \\
\midrule
\textbf{WizardCoder} & 45.24 & 41.56 & 37.46 &40.24 &20.13 &17.92 & 14.92 &16.29 \\
\textbf{WizardCoder-Wizard} & 45.25 & 40.23 & 35.36& 41.25 & 22.25 &15.31 &13.92 & 17.55 \\
\bottomrule
\end{tabular}}
\caption{Result of "Models help fine-tune themselves" experiment. we show the performance of GPT-2, GPT-Neo, WizardCoder and how they perform after they are fine-tuned by programs sampled by themselves}
\label{table:accuracy2}
\end{table*}

1: Our BTP pipeline helps code models generate better programs in the scenario where programs sampled from a better model are used to fine-tune a standard model.

2: Our BTP pipeline helps code models generate better programs in the scenario where programs sampled from the code model itself are used to fine-tune the model.

3: The best code model fine-tuned by our BTP pipeline is competitive compared to baseline methods.

4: Is there a better way to maximize the effectiveness of our BTP pipeline? (e.g., mixing sampled programs in the ER)

\subsection{Experiment Settings}
\subsubsection{Datasets} 
In recent years, a variety of open-source programming datasets have emerged, providing a robust foundation for evaluating code models. To ensure the robustness and generalizability of our proposed BTP pipeline, we applied it to fine-tune several state-of-the-art code models and evaluated them on a diverse set of popular benchmark datasets, including CodeContests from AlphaCode \cite{li2022alphacode}, APPS \cite{hendrycks2021measuring}, and HumanEval \cite{chen2021evaluating}. For HumanEval, which comprises 164 programming problems complete with function signatures, docstrings, bodies, and unit tests, we utilized all unit tests for a given problem as the test set, while the remaining descriptions were used as code generation tasks. For CodeContests, we similarly treated the problem descriptions as code generation tasks and combined all public and private test cases into a unified test set. In the case of APPS, where public and private test cases are not differentiated, we aggregated all test cases to form a comprehensive test set for each code generation task.

\subsubsection{Models} 
We categorized the models used in our experiments into two distinct groups. The first group comprises models that undergo fine-tuning, including GPT-2 and GPT-Neo. The second group consists of models employed for generating code samples. Within this latter category, we explored two scenarios: in the first, we utilized advanced code models such as GPT-4-turbo \cite{gpt42023technical}, GPT-3.5-turbo, CodeLlama-34B \cite{roziere2022unittest}, and WizardCoder-34B \cite{luo2023wizardcoder}; in the second, we utilized code models that were identical to those used for fine-tuning.

\begin{table*}[!htb]
\centering
\renewcommand{\arraystretch}{1.5} 
\scalebox{0.84}{
\begin{tabular}{lcccccccc}
\toprule
 & \multicolumn{4}{c}{\textbf{Pass Rate (\%)}} & \multicolumn{4}{c}{\textbf{Accuracy Rate (\%)}} \\
\cmidrule(lr){2-5} \cmidrule(lr){6-9}
\textbf{} & \textbf{APPS Intro.} & \textbf{APPS Inter.} & \textbf{APPS Comp.} & \textbf{APPS Mixed} & \textbf{APPS Intro.} & \textbf{APPS Inter.} & \textbf{APPS Comp.} & \textbf{APPS Mixed} \\
\midrule
\textbf{GPT-Neo-GPT4} & 51.23 & 46.39 & 38.89 & 42.12 & 26.88 & 23.34 & 17.12 & 20.54 \\
\textbf{WizardCoder} & 45.24 & 41.56 & 37.46 & 40.24 & 29.73 & 25.36 & 21.25 & 23.34 \\
\textbf{GPT-4-turbo} & 84.24 & 80.36 & 78.46 & 81.25 & 65.25 & 60.45 & 53.59 & 59.27 \\
\textbf{GPT-3.5-turbo} & 55.75 & 52.26 & 51.37 & 52.64 & 45.47 & 37.41 & 30.29 & 38.25 \\
\textbf{CodeLlama} & 53.45 & 51.26 & 49.24 & 52.27 & 28.35 & 25.92 & 23.25 & 26.78 \\
\bottomrule
\end{tabular}}
\caption{Comparison of fine-tuned code models and baseline models on the APPS dataset across different difficulty levels.}
\label{table:accuracy3}
\end{table*}

\begin{table*}[!htb]
\centering
\renewcommand{\arraystretch}{1.5} 
\scalebox{0.84}{
\begin{tabular}{lcccccccc}
\toprule
 & \multicolumn{3}{c}{\textbf{Pass Rate (\%)}} & \multicolumn{3}{c}{\textbf{Accuracy Rate (\%)}} \\
\cmidrule(lr){2-4} \cmidrule(lr){5-7}
\textbf{} & \textbf{APPS Mixed} & \textbf{CodeContests} & \textbf{HumanEval} & \textbf{APPS Mixed} & \textbf{CodeContests} & \textbf{HumanEval} \\
\midrule
\textbf{GPT-2 GPT4} & 43.27 & 35.38 & 29.25 & 5.39 & 4.25 & 4.11 \\
\textbf{CodeContests} & 36.32 & 46.93 & 38.25 & 4.82 & 6.10 & 5.25 \\
\textbf{HumanEval} & 31.92 & 25.21 & 47.93 & 2.62 & 4.25 & 8.25 \\
\textbf{Mixed} & 40.25 & 42.98 & 41.52 & 5.21 & 5.87 & 6.23 \\
\midrule
\textbf{GPT-Neo GPT4} & 45.22 & 38.18 & 30.92 & 6.48 & 5.25 & 5.59 \\
\textbf{CodeContests} & 39.11 & 48.52 & 39.26 & 5.29 & 7.20 & 6.23 \\
\textbf{HumanEval} & 33.74 & 27.93 & 50.92 & 4.92 & 6.24 & 10.29 \\
\textbf{Mixed} & 42.85 & 45.21 & 43.58 & 9.24 & 8.53 & 8.39 \\
\bottomrule
\end{tabular}}
\caption{Performance of GPT-4-turbo fine-tuned with BTP pipeline on different datasets compared with baseline models.}
\label{table:accuracy4}
\end{table*}

\subsubsection{Hyperparameter Optimization}
We conducted a series of experiments to determine the most effective hyperparameters for our models. Initially, we investigated whether beam search sampling outperforms simple sampling in terms of effectiveness. The results confirmed the superiority of beam search sampling, prompting further experiments to determine the optimal value for the beam search parameter $k$. Balancing effectiveness and resource consumption, we selected $k=3$ for our primary experiments, sampling the top-3 programs based on their probabilities. Detailed results and analysis are provided in Appendix A.

Additionally, we explored the impact of the hyperparameter $\alpha$ during the PPER phase. However, our findings revealed that the optimal $\alpha$ value varies across different models and datasets. To address this, we conducted targeted experiments across various datasets to identify the best-performing $\alpha$ for each scenario. The outcomes of these experiments, along with a comprehensive analysis, are presented in Appendix B and Appendix C.

\subsection{Fine-tuning Code Models with the BTP Pipeline}
In this section, we systematically address the four key questions posed earlier by dividing our experiments into four distinct parts.

\subsubsection{Leveraging Advanced Models to Enhance Baseline Models}
To investigate our first hypothesis, denoted as C1, we conducted an experiment to test whether the performance of baseline models can be significantly improved by leveraging advanced models within our proposed BTP (Better Transformer Programming) pipeline. Specifically, we employed the APPS dataset, which is structured into three levels of difficulty: introductory, intermediate, and competition-level tasks. These tasks were designed to assess the models' capabilities across a range of programming challenges.

As described in Table \ref{table:accuracy1}, we consolidated all three sections of the APPS dataset into a single, comprehensive dataset referred to as APPS mixed. This combined dataset was used to train and evaluate the models, ensuring that they were exposed to a diverse array of task difficulties, thereby providing a robust assessment of their generalization capabilities.

For this experiment, we selected four state-of-the-art transformer-based models: GPT-4-turbo, GPT-3.5-turbo, CodeLlama-34B, and WizardCoder-34B. These models were tasked with generating sample programs, which were subsequently used to fine-tune two baseline models: GPT-2 and GPT-Neo. The fine-tuning process involved using the sample programs generated by each advanced model to create eight fine-tuned variants of the baseline models, named as follows:

We utilized four advanced transformer models to generate sample programs, which were then used to fine-tune two baseline models. Specifically, the GPT-4-turbo, GPT-3.5-turbo, CodeLlama-34B, and WizardCoder-34B models were employed to generate samples that were subsequently used to fine-tune GPT-2 and GPT-Neo. This process resulted in eight fine-tuned models: GPT-2 fine-tuned with samples from GPT-4-turbo, GPT-3.5-turbo, CodeLlama-34B, and WizardCoder-34B, respectively named GPT-2-GPT4, GPT-2-GPT3.5, GPT-2-Llama, and GPT-2-Wizard; similarly, GPT-Neo was fine-tuned with samples from these four models, resulting in GPT-Neo-GPT4, GPT-Neo-GPT3.5, GPT-Neo-Llama, and GPT-Neo-Wizard.

After the fine-tuning process, we evaluated each of these models on the three distinct sections of the APPS dataset as well as on the combined APPS mixed dataset. The objective was to assess the extent to which the fine-tuned models could improve their performance on code generation tasks of varying complexity.

The results, presented in Table \ref{table:accuracy2}, reveal a substantial improvement in the performance of the fine-tuned models compared to their original, unmodified versions. This improvement is observed consistently across all sections of the APPS dataset, which underscores the effectiveness of our BTP pipeline. By incorporating advanced models for program sampling, we significantly enhance the capabilities of baseline transformer models like GPT-2 and GPT-Neo.

These findings strongly suggest that even relatively simple models can achieve notable performance gains when they are exposed to more advanced models during the training process, thus enabling them to perform better on complex code generation tasks.

\subsubsection{Self-Improvement Through Model Self-Fine-Tuning}
In our second experiment, we aimed to validate our second hypothesis, C2, which proposes that models can improve their own performance through a self-sampling approach within the BTP pipeline. For this experiment, we once again utilized the APPS dataset, divided into introductory, intermediate, and competition-level tasks, to evaluate the models comprehensively across different levels of difficulty.

As with our first experiment, we combined the three parts of the APPS dataset into a single, unified dataset termed APPS mixed. This ensured that each model was trained and evaluated on a diverse set of tasks, providing a rigorous test of the self-fine-tuning approach.

We selected three models for this experiment: GPT-2, GPT-Neo, and WizardCoder-34B. Each of these models was used to sample programs from the APPS mixed dataset, and the sampled programs were subsequently employed to fine-tune the same models that generated them. This process effectively creates a feedback loop, allowing the models to refine their own capabilities using their generated outputs.

The resulting self-fine-tuned models were named as follows:

GPT-2, GPT-Neo, and WizardCoder-34B were each fine-tuned using programs that they sampled themselves. This self-fine-tuning process resulted in the following models: GPT-2-2, which is GPT-2 fine-tuned with its own sampled programs; GPT-Neo-Neo, which is GPT-Neo fine-tuned with its own sampled programs; and WizardCoder-Wizard, which is WizardCoder-34B fine-tuned with its own sampled programs.

We then evaluated these self-fine-tuned models on the different sections of the APPS dataset, as well as on the APPS mixed dataset, to determine the effectiveness of this self-sampling and fine-tuning approach.

The results, shown in Table 1, indicate that while the performance improvements are modest, there is a consistent positive trend across most tasks. This suggests that the BTP pipeline can indeed enhance model performance even when models are fine-tuning themselves using their generated programs. This experiment supports the notion that models can incrementally improve their capabilities through self-guided learning, highlighting the potential of self-improvement mechanisms in transformer-based models.

\subsubsection{Comparative Analysis of the Best BTP-Generated Model and Baseline Models}
Among all the fine-tuned code models generated through the BTP pipeline, GPT-Neo-GPT4 demonstrated the best overall performance, as shown in Table \ref{table:accuracy3}. To further assess the effectiveness of our approach, we conducted a comparative analysis between GPT-Neo-GPT4 and other baseline models, as presented in Table \ref{table:accuracy4}. Although GPT-Neo-GPT4 does not outperform the most advanced baseline models, it exhibits significant improvements over its original performance and narrows the performance gap with baseline models.

This comparative analysis underscores the capability of the BTP pipeline to enhance the performance of baseline models, bringing them closer to the state-of-the-art models, albeit with some remaining performance differences.

\subsubsection{Optimizing the BTP Pipeline for Maximum Effectiveness}
To address our fourth question (Q4), we explored different strategies to maximize the effectiveness of the BTP pipeline. Specifically, we conducted experiments where we sampled programs using GPT-4-turbo from four different datasets: APPS-only, CodeContests-only, HumanEval-only, and a mixture of these datasets. We then fine-tuned GPT-2 and GPT-Neo using the sampled programs within the BTP pipeline and subsequently tested these fine-tuned models on the three datasets.

The results, as depicted in Table \ref{table:accuracy4}, indicate that when mixed datasets are used for sampling programs in the BTP pipeline, the resulting fine-tuned models show improved performance across a broader range of tasks compared to models fine-tuned on single, non-mixed datasets. However, the performance of these models on a specific dataset may not reach the level of models that were fine-tuned exclusively on that dataset.

These findings suggest that varying the datasets used for program sampling is a viable strategy to enhance the overall effectiveness of the BTP pipeline. By incorporating a diverse range of tasks during the fine-tuning process, models can achieve better generalization and performance across different code generation tasks.

\section{Conclusion}
In code generation tasks, large language models (LLMs) often need to sample a large number of programs to find a completely correct one, as even a single incorrect token can lead to failure in testing. Consequently, many sampled programs are wasted.

To utilize these resources and improve efficiency, in this work, we propose a novel algorithm called the BTP pipeline, which combines beam search sampling with prioritized experience replay to fine-tune LLMs. We empirically applied our algorithm to fine-tune several LLMs and found that they showed improvement compared to previous models. We also demonstrate that our algorithm is effective not only in scenarios where programs sampled by a better code model are used to enhance a standard code model, but also in scenarios where a code model enhances itself using programs it has sampled.

Beyond improving LLM performance in code generation tasks, we believe our BTP pipeline can be beneficial for enhancing general LLMs, particularly in cases where results sampled from LLMs are difficult to pass tests. A key limitation of this work is its reliance on code tasks and corresponding test cases. Tasks with few test cases typically result in pass rates close to zero, which can hinder the effectiveness of our algorithm. In future work, we plan to explore similar test sets and expand the available test sets to address this limitation.

\newpage
\bibliography{aaai25}

\newpage
\setcounter{theorem}{0}
\setcounter{proposition}{0}
\appendix
\onecolumn

\end{document}